\pdfoutput=1

\documentclass[11pt]{article}

\usepackage[final]{acl}

\usepackage{times}
\usepackage{latexsym}

\usepackage[T1]{fontenc}

\usepackage[utf8]{inputenc}

\usepackage{microtype}

\usepackage{inconsolata}

\usepackage{graphicx}
\usepackage{xcolor}         
\usepackage{wrapfig}
\usepackage{float}
\usepackage{caption}
\usepackage{pifont}
\usepackage{adjustbox}
\usepackage{enumitem}
\usepackage{booktabs}
\usepackage{multirow}
\usepackage{amsfonts}
\usepackage{nicefrac} 

\title{MMLU-SR: A Benchmark for Stress-Testing\\Reasoning Capability of Large Language Models}

\author{
  Wentian Wang\thanks{Visiting student at Rutgers ML Lab.}\\
  USC\\
  \AND
  Sarthak Jain\\
  Rutgers\\
  \And
    Paul Kantor\\
  Rutgers \& UW-Madison\\
  \And
  Jacob Feldman\\
  Rutgers\\
  \And
  Lazaros Gallos\\
  Rutgers\\
  \And
  Hao Wang\\
  Rutgers\\
  }

\begin{document}
\maketitle
\begin{abstract}
  We propose MMLU-SR, a novel dataset designed to measure the true comprehension abilities of Large Language Models (LLMs) by challenging their performance in question-answering tasks with modified terms. We reasoned that an agent that ``truly'' understands a concept can still evaluate it when key terms are replaced by suitably defined alternate terms, and sought to differentiate such comprehension from mere text replacement. In our study, we modified standardized test questions by replacing a key term with a dummy word along with its definition. The key term could be in the context of questions, answers, or both questions and answers. 
  Notwithstanding the high scores achieved by recent popular LLMs on the MMLU leaderboard, we found a substantial reduction in model performance after such replacement, suggesting poor comprehension. This new benchmark provides a rigorous benchmark for testing true model comprehension, and poses a challenge to the broader scientific community. 
\end{abstract}

\begin{figure*}[t] 
  \centering
  \includegraphics[width=\textwidth]{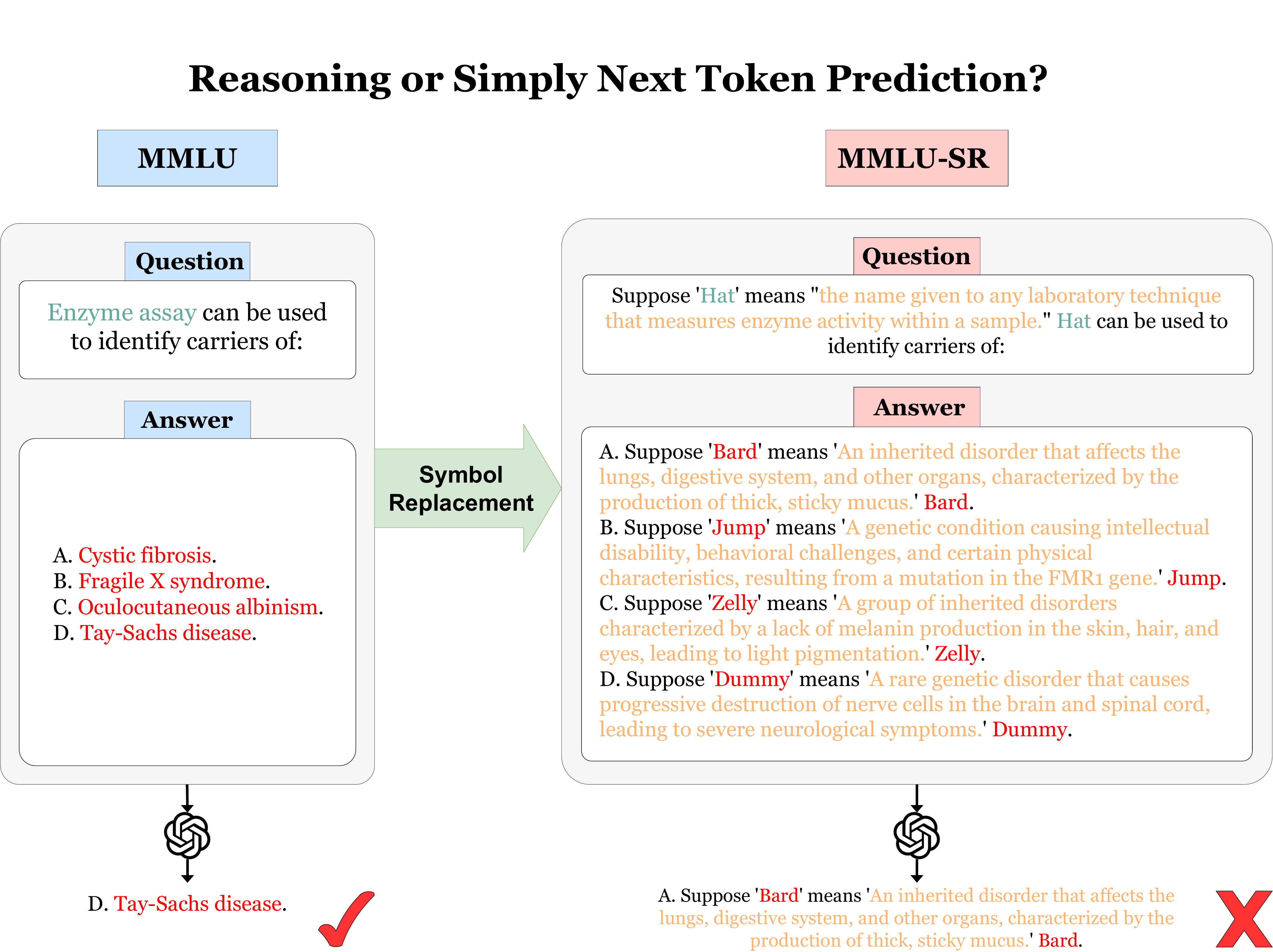} 
  \caption{Illustration of our MMLU-SR testing scenarios. The red-colored and green-colored words represent the original symbols in the MMLU dataset showing in answers and questions, which are replaced in the MMLU-SR dataset with random words followed by their definitions, shown in orange text. The example question from the MMLU dataset is correctly answered by both GPT-3.5-turbo and ChatGPT-4. However, the modified question from the MMLU-SR ``Question and Answer'' dataset is answered incorrectly by both models.}
  \label{fig:MMLUSR}
\end{figure*}
\vspace{-0.5cm}

\section{Introduction}



Large Language Models (LLMs) have achieved impressive quantitative performance on a wide range of benchmarks, natural language processing \cite{zellers2019hellaswag,wang2019glue}, general knowledge question-answering\cite{hendrycks2021mmlu,Clark2018ThinkYH}, and coding \cite{chen2021evaluating,DBLP:journals/corr/abs-2102-04664}.
Additionally, by integrating with some advanced prompting techniques, such as  Chain-of-Thought (CoT) \cite{wei2023chain} and its variants \cite{yao2023tree,trivedi2023interleaving,zhang2023automatic}, LLMs seem to exhibit a certain level of reasoning abilities including mathematics~\cite{zhang2024mario} and even causal inference/discovery~\cite{vashishtha2023causal,ICL,CausalTrans,ExposureBias}. 
However, some studies~\cite{oren2023proving} have raised concerns about data leakage (i.e., training models on the test sets), potentially rendering these results unreliable. These seemingly contradictory findings prompt the question of whether LLMs are genuinely performing reasoning tasks or merely predicting the next token. If LLMs are truly capable of reasoning, they should remain unaffected by the replacement of key symbols within the test set.

A hallmark of human intelligence is the ability to handle abstract concepts and to associate them with arbitrary terms \cite{Penn_Holyoak_2008}. With a few exceptions such as onomatopoeia, the connection between particular words and particular meanings is arbitrary, and identical concepts are invoked by different words in different human languages (e.g. \textit{dog} vs \textit{chien}). Similarly, human reasoners are capable of analogizing structural relationships from one domain to another, meaning that conceptual equivalence can be retained even when details change \cite{Gentner_Medina_1998}. It follows that true human-like comprehension should be unimpaired when terms are substituted for synonymous terms, as long as the substitution is comprehensibly defined. 

We wondered whether LLM peformance reflects true human-like comprehension in this sense, or whether it relies heavily on the specific terms used on training corpora. To assess this, we propose MMLU-SR, a new benchmark dataset that uses symbol replacement to remove some important terms from the questions and answers as shown in Figure~\ref{fig:MMLUSR}. Instead of relying on memorized terms, this approach tests whether LLMs can reason using the definitions and concepts of those terms, ensuring a more robust evaluation of their understanding. 

Our evaluations on {GPT-3.5/4, Gemini, and Llama3 families}
showed significantly lower performance on MMLU-SR compared to the original MMLU, demonstrating the effectiveness of our approach in preventing models from exploiting memorized data. MMLU-SR thus provides a more challenging and revealing test of LLMs' true reasoning abilities and understanding.

 Our findings indicate that while current LLMs excel on traditional benchmarks, they face substantial difficulties when key terms are replaced, highlighting the need for benchmarks like MMLU-SR to ensure robust and comprehensive evaluation of language models.

\section{Related Works}


\textbf{MMLU Variants Benchmarks.} 
MMLU Variants such as CMMLU \cite{li2024cmmlu} and TMMLU+ \cite{tam2024improved} are adaptations of the MMLU benchmark for non-English languages; they translate the original MMLU questions and answers into other languages, providing a way to evaluate language models' performance in non-English contexts. These benchmarks are crucial for assessing the generalizability and robustness of models across different languages and cultural settings. They preserve the original structure and intent of MMLU while enabling a broader examination of multilingual capabilities.

\textbf{Reasoning Capabilities Benchmarks.}
Several advanced reasoning benchmarks have been developed to evaluate the reasoning capabilities of language models. AGIEval \cite{zhong2023agieval} includes standardized tests and civil service exams to assess reasoning and problem-solving skills in academic and professional scenarios. BoolQ \cite{clark2019boolq} comprises over 15,000 real yes/no questions paired with Wikipedia passages to test the ability of models to infer answers from contextual information. GSM8K \cite{cobbe2021training} features 8.5K grade-school math problems requiring multistep operations, targeting the evaluation of basic to intermediate mathematical problem-solving abilities. DROP \cite{dua2019drop}, an adversarially-created reading comprehension benchmark, challenges models to navigate references and perform discrete operations such as addition and sorting, thus evaluating their capacity to understand complex texts and execute logical reasoning tasks. Beyond purely language-based evaluation, on the multimodal front, MMNeedle~\cite{MMNeedle} introduced one of the first multimodal benchmarks to evaluate long-context multimodal reasoning capabilities of multimodal LLMs. 

Unlike advanced reasoning benchmarks and MMLU variants for language extension (e.g., CMMLU and TMMLU+), 
our MMLU-SR benchmark introduces a novel approach. It enhances the challenge by replacing key words within the questions with random words, each paired with its definition, to differentiate from other benchmarks. This approach targets the models' reasoning abilities by preventing reliance on memorized terms or vocabularies. By altering key symbols, MMLU-SR ensures that the evaluation focuses on the models' understanding and reasoning, rather than their recognition of specific vocabulary, thus providing a more robust assessment of their true cognitive capabilities. We build our benchmark on the MMLU dataset because it encompasses a wide range of subjects across various domains, including Humanities, Social Sciences, STEM, and Other fields. This diverse subject matter ensures a comprehensive evaluation of language models' reasoning capabilities, in contrast to other reasoning benchmarks that often focus exclusively on specific STEM subjects.

\section{MMLU-SR Dataset}







\begin{figure}[t]  
  \centering
  \includegraphics[width=0.5\textwidth]{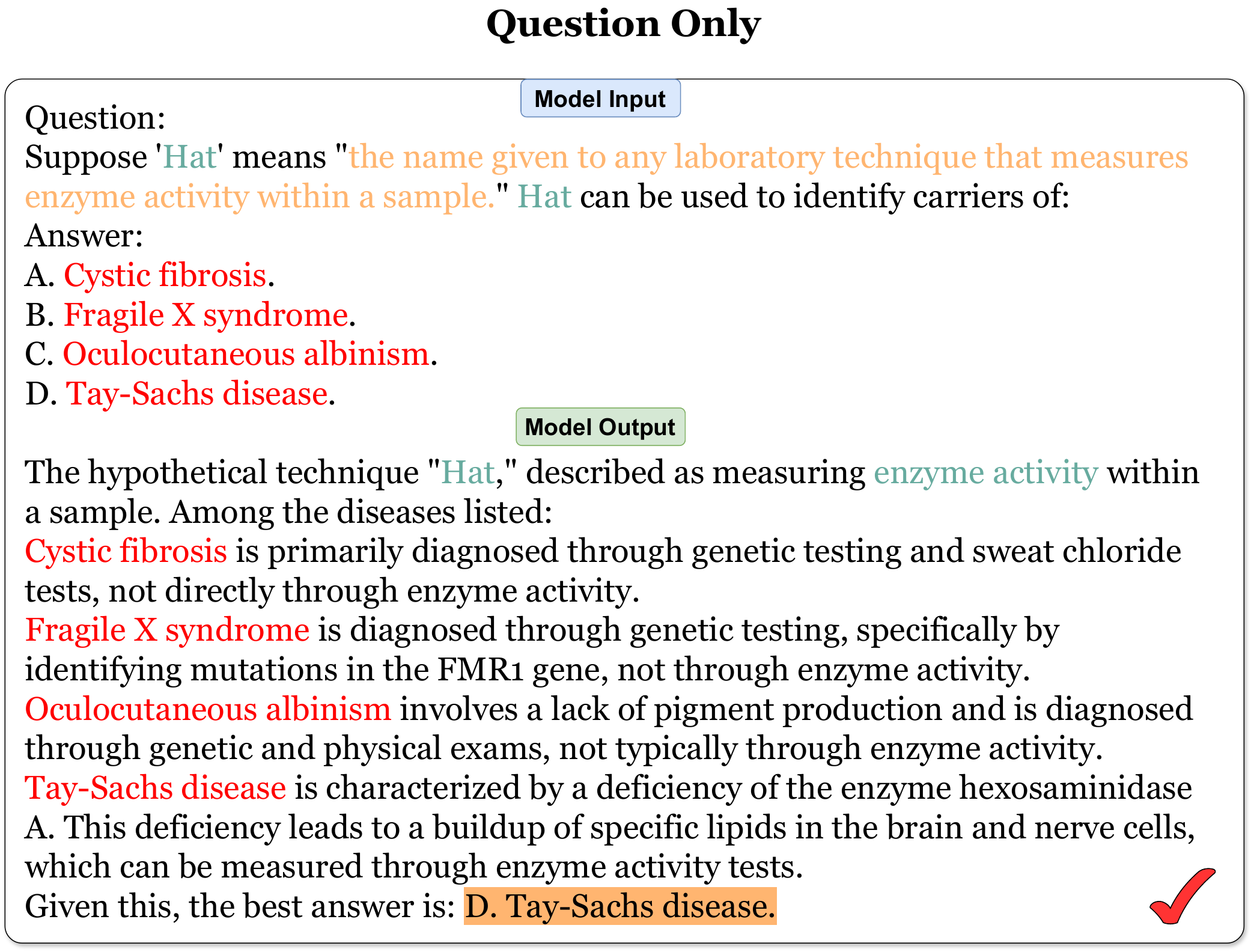}
  \caption{Example ChatGPT-4 output of MMLU-SR `Question Only''.}
  \label{fig:Question Only}
\end{figure}

\subsection{Dataset Construction}

We have developed the MMLU-SR benchmark to rigorously evaluate the reasoning and understanding capabilities of LLMs. Inspired by ObjectNet \cite{borji2020objectnet}, our benchmark contains three subsets: ``Question Only'', ``Answer Only'', and ``Question and Answer'', each offering a unique perspective on the data to comprehensively assess LLM performance. To reduce human efforts in some redundant tasks, we proposed an automatic process to generate our dataset.
\begin{enumerate}
    \item \textbf{Term Extraction and Definition Generation}: 
    We extracted key terms from the questions and answers across all 57 subjects using the assistance of \texttt{gpt-3.5-turbo}. The process involved careful few-shot prompting, and we separately extracted the contexts of questions or answers alone to ensure the model focused on extracting terms rather than solving the questions. We also retrieved appropriate definitions within the specific subject for each extracted term. For terms where the automated process provided irrelevant or inaccurate definitions, \textcolor{black}{we manually reviewed and corrected these entries (see Appendix~\ref{sec:Termapp} for details on the extent of manual modifications).}

    \item \textbf{Dictionary Creation}: 
    Once the terms and their definitions were extracted for each subject, we created JSON files where the terms served as keys and the definitions as values. This dictionary served as the basis for replacing terms in the questions and answers.

    \item \textbf{Data Replacement}: 
    Using the created dictionaries, we replaced the key terms in the questions with random dummy words followed by their definitions to create the ``Question Only'' dataset. Similarly, we did this for the answers to form the ``Answer Only'' dataset. This ensured that the context remained human-readable but required reasoning to infer the replaced terms.
    Some definitions and replacements required manual adjustments to ensure clarity and accuracy.

    \item \textbf{Combining Question and Answer Sets}: 
    After creating the ``Question Only'' and ``Answer Only'' datasets, we combined them to form the ``Question and Answer'' dataset. This step involved ensuring that the terms were consistently replaced across both questions and answers, maintaining the coherence of the dataset.

    \item \textbf{Final Adjustments}: 
    All CSV sheets were encoded in UTF-8 without headers. We manually fixed any typos that existed in the original MMLU dataset to ensure the quality and readability of the MMLU-SR dataset.
\end{enumerate}

\begin{figure}[t]  
  \centering
  \includegraphics[width=0.50\textwidth]{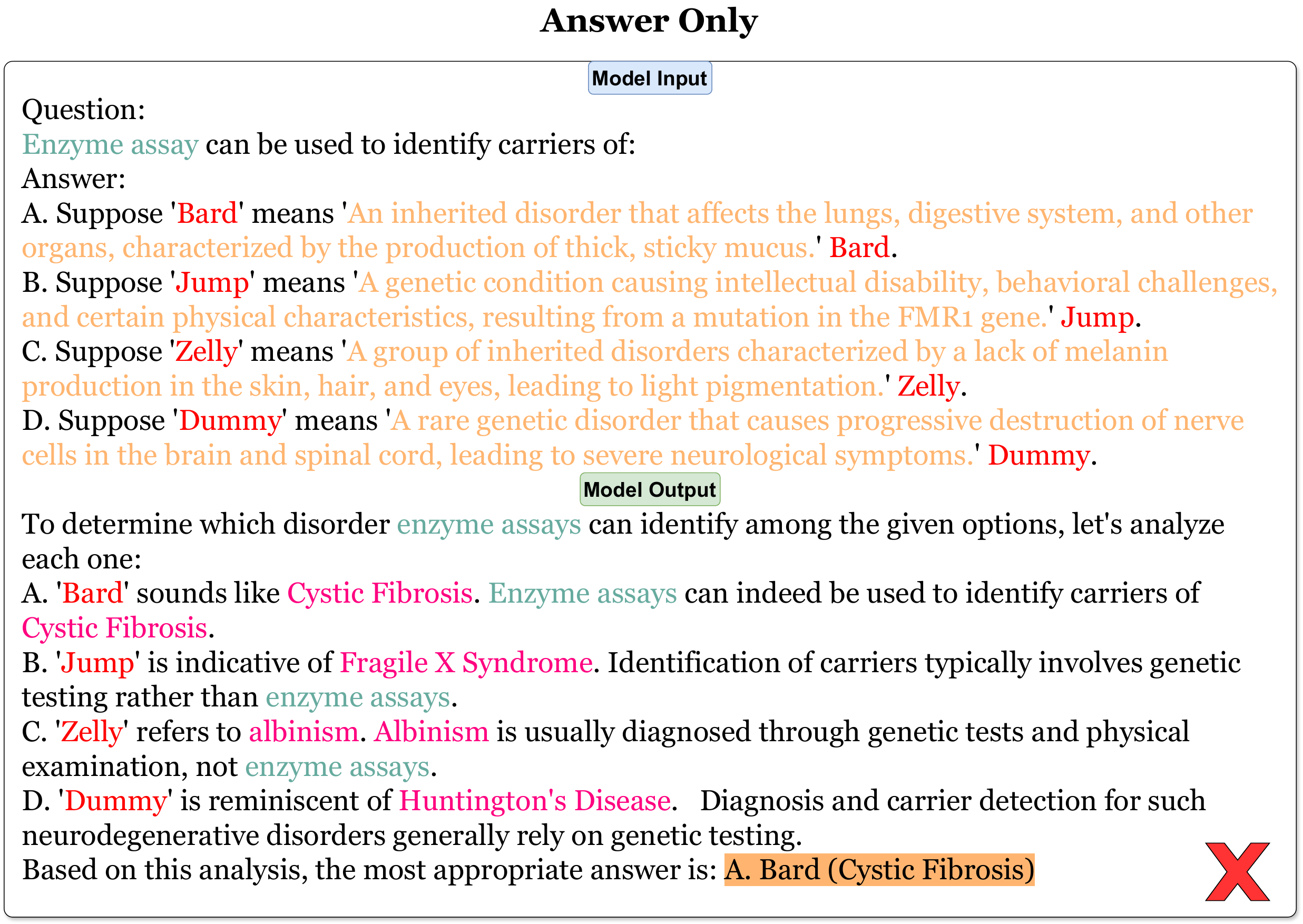} 
  \caption{Example ChatGPT-4 output of MMLU-SR ``Answer Only''.}
  \label{fig:Answer Only}
\end{figure}

The MMLU-SR dataset was created using these meticulous steps. We formed both development and test sets, with the development set used for few-shot learning and the test set reserved for evaluation. This structured approach ensured that the dataset effectively tested the reasoning abilities of LLMs, differentiating between simple pattern recognition and genuine understanding. To effectively demonstrate how our MMLU-SR dataset can challenge more sophisticated models, we use Figure~\ref{fig:Question Only}, Figure~\ref{fig:Answer Only}, and Figure~\ref{fig:Question and Answer} that feature responses from ChatGPT-4. For comparative insights, example responses of ChatGPT-3.5 are available from Table~\ref{tab:gpt3.5q}, Table~\ref{tab:gpt3.5a}, and Table~\ref{tab:gpt3.5qna} in Appendix~\ref{sec:gpt3.5}.

\subsection{``Question Only'' Dataset}

Our ``Question Only'' dataset replaces key symbols with dummy words in most questions from the original MMLU dataset and keeps the answer choices unchanged. However, some straightforward questions, such as those involving only mathematical operations and numbers or simple questions like ``Which of the following statements is true?'', remain unmodified. Generally, we changed at least one important term in the context, replacing it with a random dummy word followed by its definition; sometimes, multiple terms are replaced in the question to further test the model's reasoning capabilities.

Figure~\ref{fig:Question Only} illustrates an example where the dummy word `Hat' is used to replace term `enzyme assay' and is defined as ``the name given to any laboratory technique that measures enzyme activity within a sample.'' In this case, ChatGPT-4 successfully recognized `Hat' and understood that the critical aspect was whether the diseases listed in the answer choices could be identified through enzyme activity. The model systematically eliminated each answer option, except for the last one, as it was the only choice that could be measured through enzyme activity tests. This setup effectively tests the model’s ability to comprehend and reason based on the provided definitions, rather than relying on pre-trained knowledge of the term `enzyme assay'.

\begin{table*}[!t]
\small   
\centering
\caption{Performance of \texttt{gpt-4o-mini},\texttt{gpt-4o}, \texttt{gemini-1.5-pro}, and \texttt{llama3-70b}.}
\vspace{0.2cm}
\label{tab:results}
\begin{tabular}{@{}lcccccc@{}}
\toprule
\textbf{Dataset} & \textbf{Humanities} & \textbf{Social Sciences} & \textbf{STEM} & \textbf{Other} & \textbf{Average}  \\ \midrule
\multicolumn{6}{c}{\textbf{GPT-4o-mini}} \\ \midrule
MMLU (5-shot)                & 0.793 & 0.858 & \textbf{0.689} & 0.782 & 0.771\\
Question Only (5-shot)       & 0.744 & 0.792 & \textbf{0.621} & 0.724 & 0.710\\
Answer Only (5-shot)         & 0.659 & 0.738 & \textbf{0.602} & 0.651 & 0.655\\
Question and Answer (5-shot) & 0.588 & 0.666 & \textbf{0.531} & 0.585 & 0.585\\ \midrule
\multicolumn{6}{c}{\textbf{GPT-4o}} \\ \midrule
MMLU (5-shot)                & 0.880 & 0.906 & \textbf{0.771} & 0.854 & 0.845\\
Question Only (5-shot)       & 0.838 & 0.856 & \textbf{0.702} & 0.811 & 0.792\\
Answer Only (5-shot)         & 0.764 & 0.824 & \textbf{0.705} & 0.760 & 0.757\\
Question and Answer (5-shot) & 0.708 & 0.754 & \textbf{0.635} & 0.712 & 0.695\\ \midrule
\multicolumn{6}{c}{\textbf{Gemini-1.5-pro}} \\ \midrule
MMLU (5-shot)                &  0.849 & 0.881 & \textbf{0.802} & 0.815 & 0.832\\
Question Only (5-shot)       & 0.795 & 0.836 & \textbf{0.700} & 0.754 & 0.764\\
Answer Only (5-shot)         & 0.741 & 0.816 & 0.747 & \textbf{0.739} & 0.758\\
Question and Answer (5-shot) & 0.690 & 0.752 & \textbf{0.670} & 0.681 & 0.694\\ \midrule
\multicolumn{6}{c}{\textbf{Llama3-70B}} \\ \midrule
MMLU (5-shot)                & \textbf{0.681} & 0.868 & 0.697 & 0.814  & 0.765\\
Question Only (5-shot)       & 0.635 & 0.812 & \textbf{0.631} & 0.770 & 0.712\\
Answer Only (5-shot)         & \textbf{0.539} & 0.683 & 0.565 & 0.622 & 0.602\\
Question and Answer (5-shot) & \textbf{0.523} & 0.653 & 0.536 & 0.591  & 0.576\\ \bottomrule
\end{tabular}
\end{table*}

\begin{table*}[!t]
\footnotesize 
\centering
\caption{Relative percentage drop of accuracy in MMLU-SR compared to MMLU.}
\vspace{0.2cm}
\label{tab:drops}
\begin{tabular}{@{}lccccc@{}}
\toprule
\textbf{Dataset} & \textbf{Humanities} & \textbf{Social Sciences} & \textbf{STEM} & \textbf{Other} & \textbf{Average} \\ \midrule
\multicolumn{6}{c}{\textbf{GPT-4o-mini}} \\ \midrule
Question Only (5-shot)       & 6.18\%  & 7.69\%  & \textbf{9.87}\%  & 7.42\% & 7.91\%  \\
Answer Only (5-shot)         & \textbf{16.90\%} & 13.99\% & 12.63\% & 16.75\% & 15.05\% \\
Question and Answer (5-shot) & \textbf{25.85}\% & 22.38\% & 22.93\% & 25.19\% & 24.12\% \\ \midrule
\multicolumn{6}{c}{\textbf{GPT-4o}} \\ \midrule
Question Only (5-shot)       & 4.77\%  & 5.52\%  & \textbf{8.95}\%  & 5.03\% & 6.27\%  \\
Answer Only (5-shot)         & \textbf{13.18\%} & 9.05\% & 8.56\% & 11.01\% & 10.41\% \\
Question and Answer (5-shot) & \textbf{19.55}\% & 16.78\% & 17.64\% & 16.63\% & 17.75\% \\ \midrule
\multicolumn{6}{c}{\textbf{Gemini-1.5-pro}} \\ \midrule
Question Only (5-shot)       & 6.36\%  & 5.11\%  & \textbf{12.72\%}  & 7.48\%  & 8.17\%  \\
Answer Only (5-shot)         & \textbf{12.72\%} & 7.38\% & 6.86\% & 9.33\% & 8.89\% \\
Question and Answer (5-shot) & \textbf{18.73\%} & 14.64\% & 16.46\% & 16.44\% & 16.59\% \\ \midrule
\multicolumn{6}{c}{\textbf{Llama3-70B}} \\ \midrule
Question Only (5-shot)       & 6.75\%  & 6.45\%  & \textbf{9.47\%}  & 5.41\%  & 6.93\%  \\
Answer Only (5-shot)         & 20.85\% & 21.31\% & 18.94\% & \textbf{23.59\%} & 21.31\% \\
Question and Answer (5-shot) & 23.20\% & 24.77\% & 23.10\% & \textbf{27.40\%} & 24.71\% \\  \bottomrule
\end{tabular}
\end{table*}

\subsection{``Answer Only'' Dataset}

\begin{figure}[t]  
  \centering
  \includegraphics[width=0.50\textwidth]{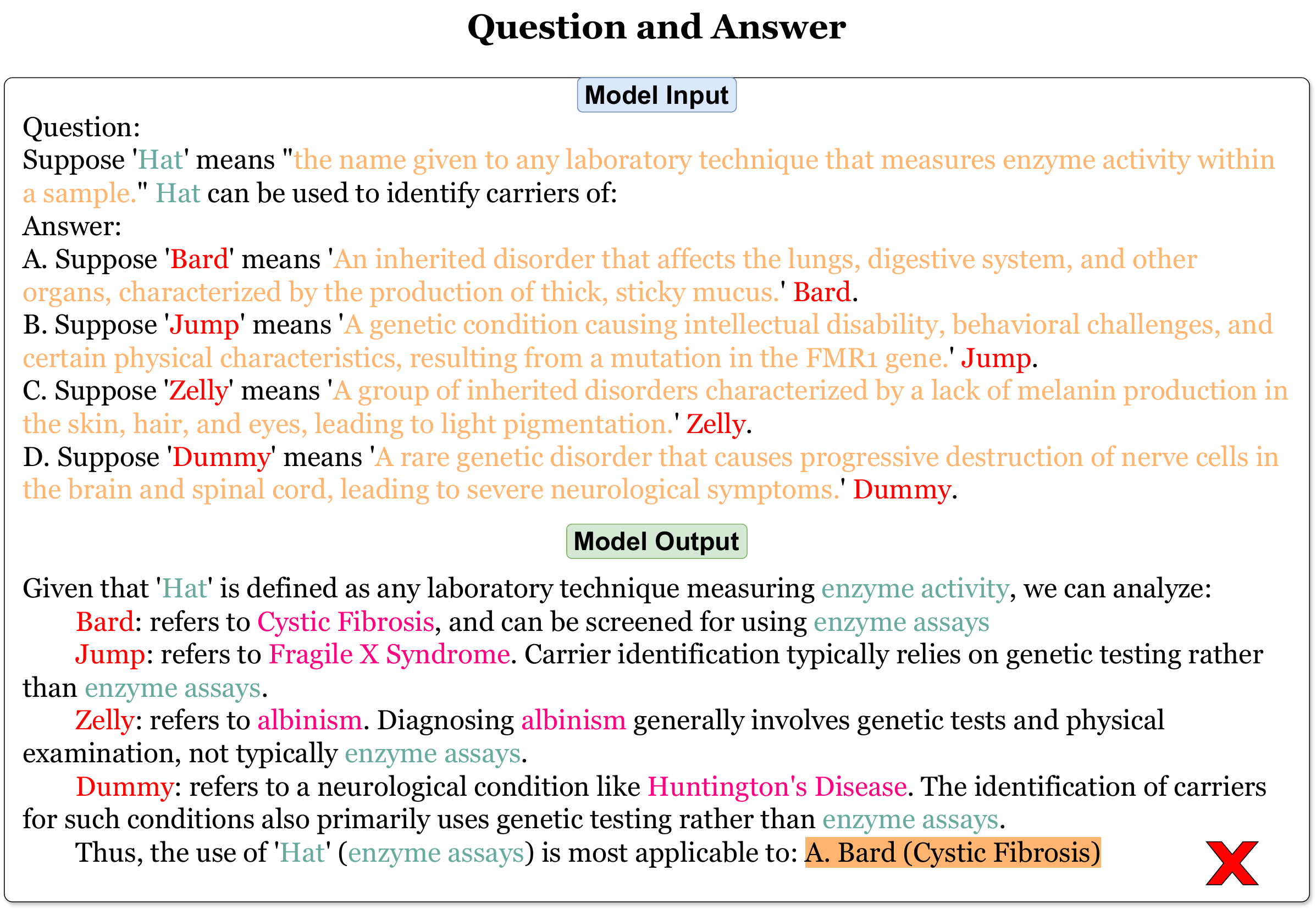}
  \caption{Example ChatGPT-4 output of MMLU-SR ``Question and Answer''.}
  \label{fig:Question and Answer}
\end{figure}
Our ``Answer Only'' dataset modifies the answer choices in most text from the original MMLU dataset similarly to the ``Question Only'' dataset and keeps the questions unchanged. We ensure that at least one answer choice is modified among all four choices. In specific cases, such as ``moral scenarios'' where the answers are simple TRUE or FALSE choices, we make substitutions like ``Suppose `Jack' means `True'.'' and ``Suppose `Luck' means `False'.'' Despite these modifications, Table~\ref{tab:Humanities} still shows the accuracy drops among all models compared to the original ``moral scenarios'' sheet from MMLU. This underscores the challenge posed to models in adapting to these symbolic substitutions.
Figure~\ref{fig:Answer Only} demonstrates that ChatGPT-4 was able to recognize the replaced terms in answer choices A, B, and C, identifying `Bard' as `Cystic Fibrosis', `Jump' as `Fragile X Syndrome', and `Zelly' as `Albinism'. The model incorrectly identified the term `Dummy' as `Huntington's Disease', while the correct term is `Tay-Sachs Disease'. 
Both disorders are indeed genetic, but they are distinct in their genetic causes and manifestations. It appears that ChatGPT-4, focusing on the broader category of `genetic disorder' from the provided definition, inadvertently linked the description to the wrong disease. 
Such misidentification led the model to persist in incorrectly affirming that choice A (`Bard' as `Cystic Fibrosis') was the correct answer (it is not).

\begin{table*}[!t]
\vspace{-10pt}
  \caption{Detailed accuracy for different Humanities subjects across different models.}
  \vspace{0.2cm}
  \label{tab:Humanities}
  \centering
  \setlength{\tabcolsep}{2pt}
  \footnotesize    
  \begin{adjustbox}{max width=\textwidth}
  \begin{tabular}{lcccccccccccc}
    \toprule
    \multirow{2}{*}{Subject} & \multicolumn{3}{c}{MMLU} & \multicolumn{3}{c}{Question Only} & \multicolumn{3}{c}{Answer Only} & \multicolumn{3}{c}{Question and Answer} \\  
     & GPT & Gemini & Llama3 & GPT & Gemini & Llama3 & GPT & Gemini & Llama3 & GPT & Gemini & Llama3  \\
     \midrule 
     Formal Logic & 0.730 & 0.698 & \textbf{0.532} & 0.603 & 0.500 & \textbf{0.484} & 0.643 & 0.579 & \textbf{0.516} & 0.556 & 0.500 & \textbf{0.460} \\
    Logical Fallacies & 0.902 & 0.902 & \textbf{0.853} & 0.883 & 0.834 & \textbf{0.810} & 0.853 & 0.847 & \textbf{0.663} & 0.834 & 0.841 & \textbf{0.564} \\
    Moral Disputes & 0.882 & \textbf{0.832} & 0.847 & 0.832 & 0.806 & \textbf{0.769} & 0.777 & 0.830 & \textbf{0.630} & 0.711 & 0.749 & \textbf{0.653} \\
    Moral Scenarios & 0.813 & 0.760 & \textbf{0.318} & 0.830 & 0.774 & \textbf{0.289} & \textbf{0.143} & 0.199 & 0.318 & 0.177 & \textbf{0.167} & 0.253 \\
    Philosophy & 0.891 & 0.865 & \textbf{0.865} & 0.778 & \textbf{0.724} & 0.772 & 0.698 & 0.756 & \textbf{0.598} & 0.582 & 0.611 & \textbf{0.582} \\
    World Religions & 0.901 & \textbf{0.895} & 0.906 & 0.895 & \textbf{0.836} & 0.895 & 0.842 & 0.813 & \textbf{0.696} & 0.825 & 0.772 & \textbf{0.684} \\
    High School European History & 0.903 & 0.885 & \textbf{0.848} & 0.885 & 0.855 & \textbf{0.830} & 0.897 & 0.849 & \textbf{0.721} & 0.861 & 0.818 & \textbf{0.739} \\
    High School Us History & 0.946 & \textbf{0.922} & 0.946 & 0.917 & 0.902 & \textbf{0.887} & 0.897 & 0.863 & \textbf{0.799} & 0.863 & 0.819 & \textbf{0.799} \\
    High School World History & 0.937 & \textbf{0.920} & 0.945 & 0.924 & 0.920 & \textbf{0.916} & 0.907 & 0.865 & \textbf{0.806} & 0.882 & \textbf{0.827} & 0.840 \\ 
    Prehistory & 0.948 & \textbf{0.901} & 0.910 & 0.904 & 0.836 & \textbf{0.793} & 0.843 & 0.803 & \textbf{0.670} & 0.790 & 0.769 & \textbf{0.670} \\
    International Law & 0.942 & 0.926 & \textbf{0.868} & 0.901 & \textbf{0.860} & 0.868 & 0.934 & 0.843 & \textbf{0.769} & 0.835 & 0.802 & \textbf{0.760} \\
    Jurisprudence & 0.898 & 0.861 & \textbf{0.852} & 0.852 & 0.861 & \textbf{0.806} & 0.861 & 0.806 & \textbf{0.602} & 0.722 & 0.750 & \textbf{0.556} \\  
    Professional Law & 0.749 & 0.666 & \textbf{0.616} & 0.683 & 0.627 & \textbf{0.583} & 0.641 & 0.585 & \textbf{0.461} & 0.563 & 0.544 & \textbf{0.461} \\ 
    \bottomrule
  \end{tabular}
  \end{adjustbox}
\end{table*}

\begin{table*}[!t]
\vspace{10pt}
  \caption{Detailed accuracy for different Social Science subjects across different models.}
  \label{tab:SocialScience}
  \vspace{0.2cm}
  \centering
  \setlength{\tabcolsep}{2pt}
  \footnotesize   
  \begin{adjustbox}{max width=\textwidth}
  \begin{tabular}{lcccccccccccc}
    \toprule
    \multirow{2}{*}{Subject} & \multicolumn{3}{c}{MMLU} & \multicolumn{3}{c}{Question Only} & \multicolumn{3}{c}{Answer Only} & \multicolumn{3}{c}{Question and Answer} \\  
     & GPT & Gemini & Llama3 & GPT & Gemini & Llama3 & GPT & Gemini & Llama3 & GPT & Gemini & Llama3  \\
     \midrule 
      Econometrics & 0.711 & 0.702 & \textbf{0.693} & 0.588 & 0.579 & \textbf{0.570} & 0.640 & 0.614 & \textbf{0.561} & 0.535 & 0.535 & \textbf{0.421} \\
    High School Macroeconomics & 0.921 & 0.880 & \textbf{0.821} & 0.849 & 0.785 & \textbf{0.779} & 0.813 & 0.785 & \textbf{0.628} & 0.721 & 0.715 & \textbf{0.572} \\
    High School Microeconomics & 0.971 & 0.929 & \textbf{0.870} & 0.903 & 0.870 & \textbf{0.773} & 0.857 & 0.815 & \textbf{0.664} & 0.769 & 0.744 & \textbf{0.571} \\
     High School Government And Politics & 0.984 & 0.974 & \textbf{0.969} & 0.979 & 0.943 & \textbf{0.938} & 0.943 & 0.922 & \textbf{0.798} & 0.922 & 0.845 & \textbf{0.782} \\    
    Public Relations & 0.836 & \textbf{0.746} & 0.755 & 0.755 & 0.755 & \textbf{0.736} & 0.664 & 0.682 & \textbf{0.600} & 0.627 & 0.646 & \textbf{0.555} \\
    Security Studies & 0.824 & 0.841 & \textbf{0.824} & 0.788 & 0.792 & \textbf{0.767} & 0.731 & 0.796 & \textbf{0.673} & 0.633 & 0.714 & \textbf{0.624} \\
    Us Foreign Policy & 0.930 & 0.940 & \textbf{0.930} & 0.920 & 0.930 & \textbf{0.890} & 0.870 & 0.880 & \textbf{0.740} & 0.810 & 0.810 & \textbf{0.780} \\
    Human Sexuality & 0.931 & 0.893 & \textbf{0.855} & 0.924 & 0.855 & \textbf{0.840} & 0.863 & 0.847 & \textbf{0.710} & 0.802 & 0.756 & \textbf{0.756} \\
    Sociology & 0.935 & \textbf{0.891} & 0.920 & 0.900 & 0.896 & \textbf{0.841} & 0.881 & 0.881 & \textbf{0.806} & 0.831 & 0.851 & \textbf{0.786} \\
    High School Geography & 0.955 & 0.939 & \textbf{0.924} & 0.894 & 0.909 & \textbf{0.833} & 0.884 & 0.864 & \textbf{0.737} & 0.813 & 0.813 & \textbf{0.662} \\
    High School Psychology & 0.965 & 0.938 & \textbf{0.921} & 0.923 & 0.917 & \textbf{0.884} & 0.927 & 0.912 & \textbf{0.719} & 0.872 & 0.859 & \textbf{0.739} \\
    Professional Psychology & 0.908 & 0.895 & \textbf{0.845} & 0.845 & 0.801 & \textbf{0.788} & 0.817 & 0.791 & \textbf{0.627} & 0.719 & 0.737 & \textbf{0.601}\\ 
    \bottomrule
  \end{tabular}
  \end{adjustbox}
\end{table*}

\subsection{``Question and Answer'' Dataset}
Our ``Question and Answer'' dataset integrates elements from both the ``Question Only'' and ``Answer Only'' datasets, replacing fundamental terms in both the questions and answer choices with dummy words followed by their definitions. As illustrated in Figure~\ref{fig:Question and Answer}, ChatGPT-4 successfully interpreted the original terms for each replaced term in answer choices A through C. However, similar to the results seen in Figure~\ref{fig:Answer Only}, the model incorrectly recognized the term in the last answer choice D (`Dummy' for Huntington's Disease), leading to an incorrect answer. This outcome contrasts with Figure~\ref{fig:Question Only}, where ChatGPT-4 correctly answered the questions when only the questions were modified. This illustrates that as complexity in context increases, with terms being replaced in both questions and answers, the model struggles to accurately identify the correct original term, consequently leading to an incorrect answer choice.

\begin{table*}[!t]
  \caption{Detailed accuracy for different STEM subjects across different models.}
  \label{tab:STEM}
  \vspace{0.5cm}
  \centering
  \setlength{\tabcolsep}{2pt}
  \footnotesize 
  \begin{adjustbox}{max width=\textwidth}
  \begin{tabular}{lcccccccccccc}
    \toprule
    \multirow{2}{*}{Subject} & \multicolumn{3}{c}{MMLU} & \multicolumn{3}{c}{Question Only} & \multicolumn{3}{c}{Answer Only} & \multicolumn{3}{c}{Question and Answer} \\  
     & GPT & Gemini & Llama3 & GPT & Gemini & Llama3 & GPT & Gemini & Llama3 & GPT & Gemini & Llama3  \\
     \midrule 
      Abstract Algebra & 0.660 & 0.690 & \textbf{0.380} & 0.470 & 0.550 & \textbf{0.370} & 0.640 & 0.730 & \textbf{0.400} & 0.460 & 0.520 & \textbf{0.400} \\
     College Mathematics & \textbf{0.490} & 0.680 & 0.510 & \textbf{0.420} & 0.630 & 0.490 & \textbf{0.440} & 0.650 & 0.460 & \textbf{0.410} & 0.610 & 0.480 \\
     High School Statistics & 0.769 & 0.866 & \textbf{0.699} & 0.708 & 0.708 & \textbf{0.657} & 0.750 & 0.829 & \textbf{0.620} & 0.644 & 0.662 & \textbf{0.597} \\
     Elementary Mathematics & 0.735 & 0.921 & \textbf{0.606} & 0.675 & 0.786 & \textbf{0.521} & 0.706 & 0.900 & \textbf{0.561} & 0.661 & 0.825 & \textbf{0.497} \\
      High School Mathematics & 0.541 & 0.700 & \textbf{0.422} & 0.537 & 0.504 & \textbf{0.356} & 0.541 & 0.615 & \textbf{0.426} & 0.511 & 0.526 & \textbf{0.367} \\
      Astronomy & 0.947 & \textbf{0.901} & 0.921 & 0.908 & \textbf{0.829} & 0.849 & 0.888 & 0.849 & \textbf{0.697} & 0.855 & 0.796 & \textbf{0.684} \\
    College Physics & 0.686 & 0.716 & \textbf{0.559} & 0.559 & 0.647 & \textbf{0.451} & 0.618 & 0.745 & \textbf{0.431} & 0.480 & 0.608 & \textbf{0.422} \\
    Conceptual Physics & 0.911 & 0.932 & \textbf{0.783} & 0.804 & 0.757 & \textbf{0.677} & 0.791 & 0.843 & \textbf{0.494} & 0.685 & 0.698 & \textbf{0.447} \\
    High School Physics & 0.748 & 0.782 & \textbf{0.563} & 0.649 & 0.556 & \textbf{0.530} & 0.589 & 0.616 & \textbf{0.477} & 0.543 & 0.596 & \textbf{0.450} \\
    College Chemistry & \textbf{0.570} & 0.610 & 0.580 & \textbf{0.540} & 0.550 & 0.570 & 0.550 & 0.530 & \textbf{0.480} & \textbf{0.480} & 0.560 & 0.470 \\
    High School Chemistry & 0.759 & 0.788 & \textbf{0.734} & 0.709 & \textbf{0.685} & 0.631 & 0.670 & 0.680 & \textbf{0.537} & 0.586 & 0.626 & \textbf{0.468} \\
     College Biology & 0.951 & \textbf{0.868} & 0.931 & 0.938 & 0.882 & \textbf{0.854} & 0.924 & 0.861 & \textbf{0.708} & 0.833 & 0.826 & \textbf{0.625} \\
    High School Biology & 0.958 & 0.929 & \textbf{0.903} & 0.932 & 0.893 & \textbf{0.858} & 0.884 & 0.858 & \textbf{0.713} & 0.858 & 0.829 & \textbf{0.729} \\
    College Computer Science & 0.790 & 0.790 & \textbf{0.670} & 0.690 & \textbf{0.610} & 0.650 & 0.760 & 0.730 & \textbf{0.610} & 0.670 & 0.660 & \textbf{0.570} \\   
    Computer Security & 0.840 & \textbf{0.820} & 0.830 & 0.830 & 0.770 & \textbf{0.750} & 0.760 & 0.730 & \textbf{0.660} & 0.760 & \textbf{0.610} & 0.720 \\
    High School Computer Science & 0.910 & 0.920 & \textbf{0.870} & 0.860 & 0.880 & \textbf{0.790} & 0.880 & 0.910 & \textbf{0.820} & 0.850 & 0.870 & \textbf{0.740} \\
    Machine Learning & 0.777 & 0.714 & \textbf{0.652} & 0.661 & 0.643 & \textbf{0.589} & 0.643 & 0.661 & \textbf{0.527} & 0.580 & 0.580 & \textbf{0.509} \\
    Electrical Engineering & 0.841 & 0.807 & \textbf{0.745} & 0.752 & 0.724 & \textbf{0.655} & 0.655 & 0.710 & \textbf{0.510} & 0.566 & 0.655 & \textbf{0.490} \\
    \bottomrule
  \end{tabular}
  \end{adjustbox}
  \vspace{-0.5cm}
\end{table*}

\begin{table*}[!t]
  \vspace{10pt}
  \caption{Detailed accuracy for different Other subjects across different models.}
  \label{tab:Other}
  \vspace{0.2cm}
  \centering
  \setlength{\tabcolsep}{2pt}
  \footnotesize 
  \begin{adjustbox}{max width=\textwidth}
  \begin{tabular}{lcccccccccccc}
    \toprule
    \multirow{2}{*}{Subject} & \multicolumn{3}{c}{MMLU} & \multicolumn{3}{c}{Question Only} & \multicolumn{3}{c}{Answer Only} & \multicolumn{3}{c}{Question and Answer} \\  
     & GPT & Gemini & Llama3 & GPT & Gemini & Llama3 & GPT & Gemini & Llama3 & GPT & Gemini & Llama3  \\
     \midrule 
      Anatomy & 0.911 & \textbf{0.793} & 0.807 & 0.874 & 0.733 & \textbf{0.726} & 0.815 & 0.667 & \textbf{0.563} & 0.726 & 0.659 & \textbf{0.578} \\
     Clinical Knowledge & 0.898 & \textbf{0.838} & 0.849 & 0.811 & 0.785 & \textbf{0.740} & 0.796 & 0.755 & \textbf{0.638} & 0.713 & 0.709 & \textbf{0.608} \\
      College Medicine & 0.832 & 0.844 & \textbf{0.757} & 0.780 & 0.786 & \textbf{0.740} & 0.798 & 0.763 & \textbf{0.647} & 0.717 & 0.740 & \textbf{0.659} \\
       Human Aging & 0.830 & \textbf{0.807} & 0.807 & 0.794 & \textbf{0.744} & 0.758 & 0.704 & 0.740 & \textbf{0.457} & 0.632 & 0.691 & \textbf{0.471} \\
       Medical Genetics & 0.960 & 0.910 & \textbf{0.830} & 0.900 & 0.850 & \textbf{0.820} & 0.840 & 0.780 & \textbf{0.570} & 0.830 & 0.740 & \textbf{0.550} \\
       Nutrition & 0.899 & 0.876 & \textbf{0.853} & 0.863 & \textbf{0.758} & 0.804 & 0.798 & 0.784 & \textbf{0.663} & 0.699 & 0.703 & \textbf{0.647} \\
        Professional Medicine & 0.956 & \textbf{0.864} & 0.868 & 0.919 & \textbf{0.776} & 0.868 & 0.901 & 0.783 & \textbf{0.754} & 0.842 & \textbf{0.735} & 0.754 \\
        Virology & 0.578 & 0.578 & \textbf{0.536} & 0.548 & 0.506 & \textbf{0.488} & 0.524 & 0.542 & \textbf{0.452} & 0.524 & 0.494 & \textbf{0.404} \\
    Business Ethics & 0.860 & 0.850 & \textbf{0.750} & 0.890 & 0.780 & \textbf{0.720} & 0.750 & 0.670 & \textbf{0.500} & 0.710 & 0.640 & \textbf{0.480} \\
    Management & 0.913 & \textbf{0.893} & 0.913 & 0.883 & \textbf{0.816} & 0.903 & \textbf{0.757} & 0.835 & 0.728 & 0.767 & 0.767 & \textbf{0.650} \\
    Marketing & 0.949 & 0.940 & \textbf{0.923} & 0.906 & 0.927 & \textbf{0.880} & 0.838 & 0.846 & \textbf{0.615} & 0.808 & 0.803 & \textbf{0.662} \\
    Global Facts & 0.650 & 0.600 & \textbf{0.530} & 0.540 & 0.540 & \textbf{0.430} & \textbf{0.580} & 0.690 & 0.540 & 0.520 & 0.470 & \textbf{0.410} \\
    Miscellaneous & 0.955 & 0.955 & \textbf{0.903} & 0.932 & 0.877 & \textbf{0.860} & 0.861 & 0.847 & \textbf{0.692} & 0.840 & 0.791 & \textbf{0.616} \\
    Professional Accounting & 0.766 & 0.663 & \textbf{0.638} & 0.716 & 0.674 & \textbf{0.596} & 0.681 & 0.638 & \textbf{0.514} & 0.631 & 0.596 & \textbf{0.489} \\
    \bottomrule
  \end{tabular}
  \end{adjustbox}
  \vskip -0.2cm
\end{table*}

\section{Experiments}
\subsection{Evaluation Protocol} 
\textcolor{black}{We evaluated seven models across OpenAI, Gemini, Llama families:{ \texttt{gpt-3.5-turbo}, \texttt{gpt-4o-mini}, \texttt{gpt-4o}, \texttt{gemini-1.0-pro}, \texttt{gemini-1.5-pro}, \texttt{llama3-8b}, and \texttt{llama3-70b}}. The evaluation for GPT and Gemini models}
was conducted using the Gemini-benchmark pipeline \cite{akter2023gemini}. For these models, we set the temperature parameter to 0 and utilized carefully crafted prompts that required responses in the format of ``Answer: Letter of Choice.'' This approach ensures that the generated responses are directly comparable and suitable for evaluation. Additionally, both models were evaluated in the 5-shot setting, using examples from our development dataset to enhance their contextual understanding.
{Llama3}
was evaluated using the lm-evaluation-harness framework \cite{eval-harness}. This model employed a different evaluation strategy; it uses log likelihood to determine the model's responses. Consistent with the other models, 
{Llama3} also uses the same 5-shot setting, ensuring a standardized comparison across all tests. {The complete results of all seven models are available in Appendix~\ref{sec:complete experiment}.}

\subsection{Results and Analysis}

\textbf{General Trend.} 
\textcolor{black}{Table~\ref{tab:results} 
shows the accuracy of the four models {\texttt{gpt-4o-mini}, \texttt{gpt-4o}, \texttt{gemini-1.5-pro}, and \texttt{llama3-70b}}}
evaluated in both MMLU and our MMLU-SR. The data highlights how each model performs in the Humanities, Social Sciences, STEM, and Other academic fields, providing average scores for each subset. We observe consistent drop in model performance across all subsets when transitioning from the standard MMLU dataset to the more challenging MMLU-SR dataset, as evidenced by the decline in average accuracy from
{0.771 on the MMLU dataset to 0.710, 0.655, and 0.585,}
on our MMLU-SR's ``Question Only'', ``Answer Only'', and ``Question and Answer'' subsets, respectively, for the {\texttt{gpt-4o-mini}}
model. This trend of decreased performance is similarly observed in the other models. 

We observe a crucial trend in decreasing accuracy across datasets: The ``Question Only'' dataset experiences the least drop, followed by the ``Answer Only'' dataset, with the most significant decline occurring in the ``Question and Answer'' dataset. This trend can be primarily attributed to two major reasons: (1) When only the question is modified, the model retains the original answer choices, facilitating the inference of the modified question's meaning; in contrast, altering the answer choices removes this contextual aid, challenging the model's ability to correctly match the question with the appropriate answer. (2) Answer choices are typically more concise and therefore lack the extensive context found in questions; consequently, replacing terms in the answers not only introduces ambiguity but also demands more complex inferential reasoning, disrupting the model’s learned pattern-recognition strategies and resulting in a greater accuracy drop. The observations above also \emph{justify the design of our MMLU-SR} on three variants (i.e., ``Question Only'', ``Answer Only'', and ``Question and Answer''). 

\textbf{Accuracy Drop in Each Category.} 
Table~\ref{tab:drops} 
shows several aspects in the relative percentage drop of accuracy in MMLU-SR \textcolor{black}{compared to that in MMLU across different categories for {\texttt{gpt-4o-mini}, \texttt{gpt-4o}, \texttt{gemini-1.5-pro}, and \texttt{llama3-70b}:}}
\begin{enumerate}
    \item \textbf{Humanities and Social Sciences.} \textcolor{black}{For {\texttt{gpt-4o-mini}} and {\texttt{gpt-4o},}}
    the accuracy drops significantly in the Humanities category, with a slightly lower drop in Social Sciences. The {\texttt{gemini-1.5-pro}}
    shows the smallest performance decline in the Humanities and Social Science categories compared to the other two models evaluated. 
    {\texttt{llama3-70b}}
    exhibits a pattern similar to {\texttt{gpt-4o-mini},}
    with the Humanities and Social Sciences categories showing a moderate percentage drop, though slightly {higher} 
    than {\texttt{gpt-4o-mini}, in the ``Answer Only'' and ``Question and Answer'' dataset.}
     \item \textbf{STEM.} For {\texttt{gemini-1.5-pro}} and {\texttt{llama3-70b},} the STEM category shows a relatively moderate decrease in accuracy across the MMLU-SR datasets. Notably, {\texttt{gemini-1.5-pro}} experiences the highest drop of 12.72\% in the ``Question Only'' dataset, indicating some sensitivity in this area. {\texttt{llama3-70b}} demonstrates a similar trend, with the highest drop of 9.47\% in the STEM category, suggesting both models retain some robustness in STEM but are still impacted by symbol replacement. On the other hand, {\texttt{gpt-4o-mini}} experiences a higher drop in the ``Answer Only'' and ``Question and Answer'' datasets, particularly with a 22.93\% drop in the latter, highlighting its relative vulnerability in this domain compared to {\texttt{gemini-1.5-pro}} and {\texttt{llama3-70b}.}
    \item \textbf{Other.} The Other category generally shows a significant drop across all models and datasets, with the highest drops often observed in the ``Question and Answer'' dataset. {For example, \texttt{gpt-4o-mini} experiences a notable drop of 25.19\%, the highest among all categories and models, indicating a high sensitivity to contextual changes in this area. Similarly, \texttt{llama3-70b} follows closely with a 27.40\% drop, which is the highest in the Other category for this model. \texttt{gemini-1.5-pro} also shows a substantial drop of 16.44\%, though slightly less compared to the other models, suggesting that the ``Other'' category, like Humanities, might be more context-dependent and hence more susceptible to performance degradation when symbols are replaced.}
\end{enumerate}

\textbf{Detailed Accuracy Drop in Each Subject.} 
Table~\ref{tab:Humanities} shows a detailed comparison of accuracy scores across different models evaluated on various subjects in the Humanities category. \textcolor{black}{The MMLU scores serve as a baseline for comparison. {\texttt{gpt-4o}} demonstrates exceptional performance across most subjects in this category, often leading in accuracy, particularly in complex subjects like Philosophy and International Law. {\texttt{gemini-1.5-pro}} also shows strong performance, but {\texttt{gpt-4o}} frequently matches or exceeds its accuracy. Notably, {\texttt{gpt-4o}} performs particularly well in subjects like High School World History and Jurisprudence. However, all models continue to struggle with Moral Scenarios, where the accuracy score drops significantly, particularly for {\texttt{llama3-70b},} which shows a drastic decrease, reflecting a higher sensitivity to the challenges posed by the MMLU-SR datasets}


Table~\ref{tab:SocialScience} shows a detailed comparison of accuracy across different models evaluated on various subjects in the Social Science category. We observe that all models perform exceptionally well in Social Science on MMLU, particularly in High School Government and Politics, \textcolor{black}{where {\texttt{gpt-4o}} achieves an impressive accuracy of 0.984. While there is still a drop in accuracy from MMLU to MMLU-SR's {``Question and Answer'' dataset, {\texttt{gpt-4o}} demonstrates remarkable resilience, maintaining accuracy levels around 0.7$\sim$0.9 across most subjects. This performance significantly outpaces the other models, particularly in subjects like High School Psychology and Sociology.} The drop in accuracy, though less pronounced for {\texttt{gpt-4o}}, still illustrates how our symbol replacement method increases difficulty, effectively stress-testing the models' reasoning capabilities versus mere memorization of pre-trained terms.}

Table~\ref{tab:STEM} shows a detailed comparison of accuracy across various STEM subjects for different models. Each model demonstrated varying degrees of success across the subjects, with notable difficulties in some areas. \textcolor{black}{College Mathematics and High School Mathematics remain challenging for all models, including {\texttt{gpt-4o}}, with accuracy dropping to around 0.4 to 0.5 in MMLU-SR's ``Question and Answer'' dataset. However, {\texttt{gpt-4o}} shows marked improvement in subjects like Astronomy, College Biology, and High School Biology, maintaining high accuracy even in the more challenging MMLU-SR datasets.} The subject with the lowest accuracies among all models is still {High School Mathematics, where \texttt{llama3-70b} struggles the most, especially in the Answer Only'' and ``Question and Answer'' datasets.} Similarly, {College Physics and Abstract Algebra} also show significant drops in accuracy across all models, highlighting the persistent challenges in subjects involving extensive calculations and complex problem-solving.


Table~\ref{tab:Other} \textcolor{black}{shows a detailed comparison of accuracy scores across different models evaluated on various subjects in the Other category. We observe that {\texttt{gpt-4o}} performs exceptionally well in MMLU, with accuracy consistently above 0.9 in most subjects, significantly outperforming other models. {Marketing stands out with a particularly high accuracy of 0.949 for {\texttt{gpt-4o}}, indicating outstanding performance in this subject.} Professional Accounting shows improved performance with {\texttt{gpt-4o}}, achieving an accuracy of 0.766 in MMLU. {Virology} remains challenging, but {\texttt{gpt-4o}} shows improvement with an accuracy of 0.578. While there is still a drop in accuracy from MMLU to MMLU-SR's {``Question and Answer'' dataset, {\texttt{gpt-4o}} maintains relatively high performance, with accuracy generally staying above 0.7 for most subjects. Even in challenging areas like Virology and Global Facts, {\texttt{gpt-4o}} demonstrates resilience, maintaining accuracy levels significantly higher than other models.}}

\textbf{CoT and System Instruction.} 
We developed a simple baseline to test our MMLU-SR dataset on more recent and sophisticated models like GPT-4. This involves adding the instruction ``Let's think step by step'' at the end of answer choices to enable zero-shot CoT prompting. As shown in Table~\ref{tab:cot} from Appendix~\ref{sec:cot_examples},
we also included a system instruction informing ChatGPT-4 that the following questions would involve symbol replacement with arbitrary definitions. However, the example demonstrates that despite applying (zero-shot) CoT, the model still incorrectly interprets the term `Dummy' in choice D as `neurodegenerative disorder,' leading to the wrong answer, choice A. We applied this system instruction across the entire MMLU-SR dataset as well, with results shown in 
Table~\ref{tab:system} from Appendix~\ref{sec:sys_examples}.
The results indicate that while the system instruction slightly improves accuracy in the ``Question Only'' and ``Answer Only'' datasets, the model still struggles with the increased complexity in the ``Question and Answer'' dataset.

\section{Conclusion}
We introduced MMLU-SR, a novel benchmark that challenges LLMs by replacing key terms in questions with random words followed by their definitions, aiming to test the models' reasoning and comprehension abilities rather than their memorization skills. Our evaluation across multiple domains revealed that popular LLMs suffer from significant drops in performance with these modifications, highlighting their reliance on memorized terms. MMLU-SR's unique approach addresses concerns about overfitting to traditional benchmarks and provides a more rigorous measure of true language understanding. This dataset will enable researchers to better identify and address the reasoning limitations of current LLMs, fostering the development of more robust and genuinely intelligent models.

\section{Acknowledgements}
We thank the anonymous reviewers for their thoughtful feedback and for recognizing the significance and contributions of our research. This research is based on work partially supported by DARPA Grant HR00112420363, NSF Grant IIS-2127918, and NSF CAREER Award IIS-2340125. 

\bibliography{ref}

\appendix

\section{CoT Examples}
\label{sec:cot_examples}

Table~\ref{tab:cot} shows an example of incorrect answer using zero-shot CoT with a system instruction produced by the ChatGPT-4 on MMLU-SR's ``Question and Answer'' dataset. The correct answer is choice D, but ChatGPT-4 responded with choice A.

\begin{table*}[!t]
\caption{Example of incorrect output using zero-shot CoT with system instruction.}
\label{tab:cot}
\vspace{0.2cm}
\centering
\begin{tabular}{p{0.19\textwidth} p{0.75\textwidth}}
\toprule
\textbf{System Prompt} & In each of the questions that I ask, I will replace some of the words that you might know with a word that is arbitrarily assigned a specific meaning just for this test.  The meaning of these arbitrary definition may change with every question. 
\vspace{-0.4cm}

\\
\midrule \textbf{Input} & Suppose `Hat' means `the name given to any laboratory technique that measures enzyme activity within a sample.' Hat can be used to identify carriers of:	\vspace{0.2cm}\\ &
A. Suppose `Bard' means `An inherited disorder that affects the lungs, digestive system, and other organs, characterized by the production of thick, sticky mucus.' Bard.		\vspace{0.2cm}\\
& B. Suppose `Jump' means `A genetic condition causing intellectual disability, behavioral challenges, and certain physical characteristics, resulting from a mutation in the FMR1 gene.' Jump.	\vspace{0.2cm}\\
& C. Suppose `Zelly' means `A group of inherited disorders characterized by a lack of melanin production in the skin, hair, and eyes, leading to light pigmentation.' Zelly.		\vspace{0.2cm}\\ 
& D. Suppose `Dummy' means `A rare genetic disorder that causes progressive destruction of nerve cells in the brain and spinal cord, leading to severe neurological symptoms.' Dummy.\vspace{0.2cm}\\ 
&

Let's think step by step. 
\vspace{-0.2cm}
\\ \midrule\textbf{Output} 
 & In this scenario, Hat is used to detect variations or abnormalities in enzyme levels that might be indicative of certain genetic disorders.
\vspace{0.2cm}\\ & A: Bard refers to cystic fibrosis. Areas of land (Jack) often have boundaries (Winky) that separate them from other areas.
\vspace{0.2cm}\\ & B: Jump refers to Fragile X syndrome,  primarily involves genetic testing to detect the specific mutation rather than enzyme activity measurements.
\vspace{0.2cm}\\ & C: Zelly refers to conditions such as albinism, which are typically diagnosed based on pigment absence rather than enzyme activity, although some biochemical tests can be relevant in specific contexts.
\vspace{0.2cm}\\ & D: Dummy could be referring to several neurodegenerative disorders, which might be associated with enzyme activities related to nerve cell function or metabolic processes, but generally, such disorders are diagnosed through genetic markers and neuroimaging rather than enzyme activity alone.
\vspace{0.2cm}\\ & Given these explanations, the best match for `Hat' in terms of being a technique to measure enzyme activity useful for diagnosis would be A, `Bard' as cystic fibrosis, due to its direct relationship with enzyme dysfunction and measurement. \ding{55}\\
\bottomrule
\end{tabular}
\end{table*}

\section{Examples of ChatGPT-3.5 Response}
\label{sec:gpt3.5}
Table~\ref{tab:gpt3.5q} shows an example of incorrect answer produced by the ChatGPT-3.5 on MMLU-SR's ``Question and Answer'' dataset. 
Table~\ref{tab:gpt3.5a} shows an example of incorrect answer produced by the ChatGPT-3.5 on MMLU-SR's ``Question and Answer'' dataset. 
Table~\ref{tab:gpt3.5qna} shows an example of incorrect answer produced by the ChatGPT-3.5 on MMLU-SR's ``Question and Answer'' dataset. 

\begin{table*}[!t]
\caption{Example of incorrect response generated by ChatGPT-3.5 in ``Question Only'' dataset.}
\label{tab:gpt3.5q}
\vspace{0.2cm}
\centering
\begin{tabular}{p{0.19\textwidth} p{0.75\textwidth}}
\toprule
\textbf{Input} & Suppose `Hat' means ``the name given to any laboratory technique that measures enzyme activity within a sample.'' Hat can be used to identify carriers of:	\vspace{0.2cm}\\ &
A. Cystic fibrosis.		\vspace{0.2cm}\\
& B. Fragile X syndrome.	\vspace{0.2cm}\\
& C. Oculocutaneous albinism.		\vspace{0.2cm}\\ 
& D. Tay-Sachs disease
\vspace{-0.2cm}
\\ \midrule\textbf{Output} 
 &  ``Hat'' in this context refers to an enzyme assay technique rather than a genetic condition or disease. Therefore, it wouldn't be used to identify carriers of genetic disorders like cystic fibrosis, Fragile X syndrome, oculocutaneous albinism, or Tay-Sachs disease. It's essential to keep in mind the specific context in which terms are used in scientific fields! \ding{55}\\
\bottomrule
\end{tabular}
\end{table*}

\begin{table*}[!t]
\caption{Example of incorrect response generated by ChatGPT-3.5 in ``Answer Only'' dataset.}
\label{tab:gpt3.5a}
\vspace{0.2cm}
\centering
\begin{tabular}{p{0.19\textwidth} p{0.75\textwidth}}
\toprule
\textbf{Input} & Enzyme assay can be used to identify carriers of:	\vspace{0.2cm}\\ &
A. Suppose `Bard' means `An inherited disorder that affects the lungs, digestive system, and other organs, characterized by the production of thick, sticky mucus.' Bard.		\vspace{0.2cm}\\
& B. Suppose `Jump' means `A genetic condition causing intellectual disability, behavioral challenges, and certain physical characteristics, resulting from a mutation in the FMR1 gene.' Jump.	\vspace{0.2cm}\\
& C. Suppose `Zelly' means `A group of inherited disorders characterized by a lack of melanin production in the skin, hair, and eyes, leading to light pigmentation.' Zelly.		\vspace{0.2cm}\\ 
& D. Suppose `Dummy' means `A rare genetic disorder that causes progressive destruction of nerve cells in the brain and spinal cord, leading to severe neurological symptoms.' Dummy.
\vspace{0.2cm}
\\ \midrule\textbf{Output} 
 &  An enzyme assay could be used to identify carriers of Bard, Jump, Zelly, or Dummy if the specific enzyme associated with each disorder is known and can be tested for. Enzyme assays are commonly used in genetic testing to detect the presence or absence of specific enzymes or their activity levels, which can provide information about an individual's genetic makeup and potential carrier status for certain genetic disorders. \ding{55}\\
\bottomrule
\end{tabular}
\end{table*}

\begin{table*}[!t]
\caption{Example of incorrect response generated by ChatGPT-3.5 in ``Question and Answer'' dataset.}
\label{tab:gpt3.5qna}
\vspace{0.2cm}
\centering
\begin{tabular}{p{0.19\textwidth} p{0.75\textwidth}}
\toprule
\textbf{Input} & Suppose 'Hat' means ``the name given to any laboratory technique that measures enzyme activity within a sample.'' Hat can be used to identify carriers of	\vspace{0.2cm}\\ &
A. Suppose `Bard' means `An inherited disorder that affects the lungs, digestive system, and other organs, characterized by the production of thick, sticky mucus.' Bard.		\vspace{0.2cm}\\
& B. Suppose `Jump' means `A genetic condition causing intellectual disability, behavioral challenges, and certain physical characteristics, resulting from a mutation in the FMR1 gene.' Jump.	\vspace{0.2cm}\\
& C. Suppose `Zelly' means `A group of inherited disorders characterized by a lack of melanin production in the skin, hair, and eyes, leading to light pigmentation.' Zelly.		\vspace{0.2cm}\\ 
& D. Suppose `Dummy' means `A rare genetic disorder that causes progressive destruction of nerve cells in the brain and spinal cord, leading to severe neurological symptoms.' Dummy.

\vspace{-0.2cm}
\\ \midrule\textbf{Output} 
 & Based on the provided definitions: \vspace{0.2cm}\\
& A. Hat (enzyme activity measurement) can be used to identify carriers of Zelly (lack of melanin production disorders). \vspace{0.2cm}\\
& Explanation: Enzyme activity measurements are often used in medical diagnostics to identify various conditions, including genetic disorders. In this case, the lack of melanin production, as seen in Zelly disorders, can be identified using enzyme activity tests. \ding{55}\\
\bottomrule
\end{tabular}
\end{table*}

\section{System Instruction Example}
Table~\ref{tab:system} shows the performance comparison of \texttt{gemini-1.0-pro} with and without using the system instruction ``In each of the questions that I ask, I will replace some of the words that you might know with a word that is arbitrarily assigned a specific meaning just for this test.  The meaning of these arbitrary definition may change with every question.''
\label{sec:sys_examples}

\begin{table*}[!t]
\centering
\small
\caption{Performance comparison of the \texttt{gemini-1.0-pro} model with and without the system instruction.}
\label{tab:system}
\vspace{0.2cm}
\begin{tabular}{@{}lccccc@{}}
\toprule
\textbf{Dataset} & \textbf{Humanities} & \textbf{Social Sciences} & \textbf{STEM} & \textbf{Other} & \textbf{Average} \\ \midrule
\multicolumn{6}{c}{\textbf{Gemini-1.0-pro}} \\ \midrule
Question Only (5-shot)       & 0.687 & 0.744 & \textbf{0.539} & 0.658 & 0.645\\
Answer Only (5-shot)         & 0.619 & 0.670 & \textbf{0.504} & 0.591 & 0.586\\
Question and Answer (5-shot) & 0.582 & 0.622 & \textbf{0.472} & 0.544 & 0.546\\ \midrule
\multicolumn{6}{c}{\textbf{Gemini-1.0-pro with System Instruction}} \\ \midrule
Question Only                & 0.688 & 0.735 & \textbf{0.535} & 0.650 & 0.652 \\
Answer Only                  & 0.619 & 0.661 & \textbf{0.502} & 0.576 & 0.590 \\
Question and Answer          & 0.570 & 0.612 & \textbf{0.460} & 0.537 & 0.545 \\
\bottomrule
\end{tabular}
\end{table*}

\section{Hard MMLU-SR Example}
\label{sec:hard}

Table~\ref{tab:hard_exam} illustrates a challenging example mentioned in the ``Question and Answer'' section in the main paper. In the question, we substituted ``enzyme assay'' with a synonym, ``protein'', and used the original terms from the MMLU dataset as placeholders in the answer choices, each paired with a definition. ChatGPT-4 correctly identified the term ``Tay-Sachs Disease'' but associated it with the incorrect answer choice and definition, which was meant for ``Fragile X syndrome.'' This example demonstrates that ChatGPT-4 can be misled by the complexity of the dataset, tending to rely on pre-existing knowledge rather than engaging in substantive reasoning to align terms with their definitions.

\begin{table*}[!t]
\caption{Example of a hard version of MMLU-SR.}
\label{tab:hard_exam}
\vspace{0.2cm}
\centering
\begin{tabular}{p{0.19\textwidth} p{0.75\textwidth}}
\toprule
\textbf{Input} & Suppose `Protein' means `the name given to any laboratory technique that measures enzyme activity within a sample.' Protein can be used to identify carriers of:	\vspace{0.2cm}\\ &
A. Suppose `Oculocutaneous albini' means `An inherited disorder that affects the lungs, digestive system, and other organs, characterized by the production of thick, sticky mucus.' Oculocutaneous albini.		\vspace{0.2cm}\\
& B. Suppose `Tay-Sachs diseas' means `A genetic condition causing intellectual disability, behavioral challenges, and certain physical characteristics, resulting from a mutation in the FMR1 gene.' Tay-Sachs diseas.	\vspace{0.2cm}\\
& C. Suppose `Cystic fibrosi' means `A group of inherited disorders characterized by a lack of melanin production in the skin, hair, and eyes, leading to light pigmentation.' Cystic fibrosi.		\vspace{0.2cm}\\ 
& D. Suppose `Fragile X syndrom' means `A rare genetic disorder that causes progressive destruction of nerve cells in the brain and spinal cord, leading to severe neurological symptoms.' Fragile X syndrom.

\vspace{-0.2cm}
\\ \midrule\textbf{Output} 
 & The given problem seems to involve identifying which condition might be detected through a laboratory technique measuring enzyme activity. From the provided definitions, none of the conditions directly indicate an enzyme-related pathology except for a possible misalignment in the definitions and names: \vspace{0.2cm}\\
&

A: Oculocutaneous albini is defined incorrectly and does not relate to the production of enzymes. \vspace{0.2cm}\\
&

B: Tay-Sachs disease is a genetic disorder characterized by the destruction of nerve cells in the brain due to a deficiency of a specific enzyme (Hexosaminidase A). This is actually a plausible answer if we correct the name to Tay-Sachs disease, which fits the description of a disease involving enzyme activity. \vspace{0.2cm}\\
&

C: Cystic fibrosis and D: Fragile X syndrome are misdefined and do not relate to conditions involving measurable enzyme activity relevant to their actual conditions. \vspace{0.2cm}\\
&
Thus, the most likely answer, B. Tay-Sachs disease (with the corrected definition and name). \ding{55}\\
\bottomrule
\end{tabular}
\end{table*}

\begin{table*}[!t]
\small   
\centering
\caption{Complete performance of \texttt{gpt-3.5-turbo}, \texttt{gpt-4o-mini}, \texttt{gpt-4o}, \texttt{gemini-1.0-pro}, \texttt{gemini-1.5-pro}, \texttt{llama3-8b}, and \texttt{llama3-70b}.}
\vspace{0.2cm}
\label{tab:completeresults}
\begin{tabular}{@{}lcccccc@{}}
\toprule
\textbf{Dataset} & \textbf{Humanities} & \textbf{Social Sciences} & \textbf{STEM} & \textbf{Other} & \textbf{Average}  \\ \midrule
\multicolumn{6}{c}{\textbf{GPT-3.5-turbo}} \\ \midrule
MMLU (5-shot)                & 0.723 & 0.770 & \textbf{0.554} & 0.714 & 0.677\\
Question Only (5-shot)       & 0.661 & 0.702 & \textbf{0.506} & 0.641 & 0.616\\
Answer Only (5-shot)         & 0.540 & 0.595 & \textbf{0.441} & 0.538 & 0.520\\
Question and Answer (5-shot) & 0.469 & 0.523 & \textbf{0.396} & 0.476 & 0.459\\ \midrule
\multicolumn{6}{c}{\textbf{GPT-4o-mini}} \\ \midrule
MMLU (5-shot)                & 0.793 & 0.858 & \textbf{0.689} & 0.782 & 0.771\\
Question Only (5-shot)       & 0.744 & 0.792 & \textbf{0.621} & 0.724 & 0.710\\
Answer Only (5-shot)         & 0.659 & 0.738 & \textbf{0.602} & 0.651 & 0.655\\
Question and Answer (5-shot) & 0.588 & 0.666 & \textbf{0.531} & 0.585 & 0.585\\ \midrule
\multicolumn{6}{c}{\textbf{GPT-4o}} \\ \midrule
MMLU (5-shot)                & 0.880 & 0.906 & \textbf{0.771} & 0.854 & 0.845\\
Question Only (5-shot)       & 0.838 & 0.856 & \textbf{0.702} & 0.811 & 0.792\\
Answer Only (5-shot)         & 0.764 & 0.824 & \textbf{0.705} & 0.760 & 0.757\\
Question and Answer (5-shot) & 0.708 & 0.754 & \textbf{0.635} & 0.712 & 0.695\\ \midrule
\multicolumn{6}{c}{\textbf{Gemini-1.0-pro}} \\ \midrule
MMLU (5-shot)                & 0.728 & 0.758 & \textbf{0.596} & 0.703 & 0.686\\
Question Only (5-shot)       & 0.687 & 0.744 & \textbf{0.539} & 0.658 & 0.645\\
Answer Only (5-shot)         & 0.619 & 0.670 & \textbf{0.504} & 0.591 & 0.586\\
Question and Answer (5-shot) & 0.582 & 0.622 & \textbf{0.472} & 0.544 & 0.546\\ \midrule
\multicolumn{6}{c}{\textbf{Gemini-1.5-pro}} \\ \midrule
MMLU (5-shot)                &  0.849 & 0.881 & \textbf{0.802} & 0.815 & 0.832\\
Question Only (5-shot)       & 0.795 & 0.836 & \textbf{0.700} & 0.754 & 0.764\\
Answer Only (5-shot)         & 0.741 & 0.816 & 0.747 & \textbf{0.739} & 0.758\\
Question and Answer (5-shot) & 0.690 & 0.752 & \textbf{0.670} & 0.681 & 0.694\\ \midrule
\multicolumn{6}{c}{\textbf{Llama3-8B}} \\ \midrule
MMLU (5-shot)                & 0.593 & 0.757 & \textbf{0.557} & 0.729 & 0.651\\
Question Only (5-shot)       & 0.546 & 0.685 & \textbf{0.507} & 0.668 & 0.595\\
Answer Only (5-shot)         & \textbf{0.455} & 0.599 & 0.460 & 0.557 & 0.510\\
Question and Answer (5-shot) & \textbf{0.421} & 0.538 & 0.424 & 0.499 & 0.465\\ \midrule
\multicolumn{6}{c}{\textbf{Llama3-70B}} \\ \midrule
MMLU (5-shot)                & \textbf{0.681} & 0.868 & 0.697 & 0.814  & 0.765\\
Question Only (5-shot)       & 0.635 & 0.812 & \textbf{0.631} & 0.770 & 0.712\\
Answer Only (5-shot)         & \textbf{0.539} & 0.683 & 0.565 & 0.622 & 0.602\\
Question and Answer (5-shot) & \textbf{0.523} & 0.653 & 0.536 & 0.591  & 0.576\\
\bottomrule
\end{tabular}
\end{table*}

\begin{table*}[!t]
\footnotesize 
\centering
\caption{Complete relative percentage drop of accuracy in MMLU-SR compared to MMLU.}
\vspace{0.2cm}
\label{tab:completedrops}
\begin{tabular}{@{}lccccc@{}}
\toprule
\textbf{Dataset} & \textbf{Humanities} & \textbf{Social Sciences} & \textbf{STEM} & \textbf{Other} & \textbf{Average} \\ \midrule
\multicolumn{6}{c}{\textbf{GPT-3.5-turbo}} \\ \midrule
Question Only (5-shot)       & 8.58\%  & 8.83\%  & 8.67\%  & \textbf{10.22}\% & 9.08\%  \\
Answer Only (5-shot)         & \textbf{25.31\%} & 22.73\% & 20.40\% & 24.65\% & 23.27\% \\
Question and Answer (5-shot) & \textbf{35.12}\% & 32.08\% & 28.52\% & 33.30\% & 32.26\% \\ \midrule
\multicolumn{6}{c}{\textbf{GPT-4o-mini}} \\ \midrule
Question Only (5-shot)       & 6.18\%  & 7.69\%  & \textbf{9.87}\%  & 7.42\% & 7.91\%  \\
Answer Only (5-shot)         & \textbf{16.90\%} & 13.99\% & 12.63\% & 16.75\% & 15.05\% \\
Question and Answer (5-shot) & \textbf{25.85}\% & 22.38\% & 22.93\% & 25.19\% & 24.12\% \\ \midrule
\multicolumn{6}{c}{\textbf{GPT-4o}} \\ \midrule
Question Only (5-shot)       & 4.77\%  & 5.52\%  & \textbf{8.95}\%  & 5.03\% & 6.27\%  \\
Answer Only (5-shot)         & \textbf{13.18\%} & 9.05\% & 8.56\% & 11.01\% & 10.41\% \\
Question and Answer (5-shot) & \textbf{19.55}\% & 16.78\% & 17.64\% & 16.63\% & 17.75\% \\ \midrule
\multicolumn{6}{c}{\textbf{Gemini-1.0-pro}} \\ \midrule
Question Only (5-shot)       & 5.63\%  & 1.85\%  & \textbf{9.56\%}  & 6.40\%  & 5.86\%  \\
Answer Only (5-shot)         & 14.96\% & 11.61\% & 15.44\% & \textbf{15.91\%} & 14.48\% \\
Question and Answer (5-shot) & 20.05\% & 17.94\% & 20.81\% & \textbf{22.60\%} & 20.85\% \\ 
\midrule
\multicolumn{6}{c}{\textbf{Gemini-1.5-pro}} \\ \midrule
Question Only (5-shot)       & 6.36\%  & 5.11\%  & \textbf{12.72\%}  & 7.48\%  & 8.17\%  \\
Answer Only (5-shot)         & \textbf{12.72\%} & 7.38\% & 6.86\% & 9.33\% & 8.89\% \\
Question and Answer (5-shot) & \textbf{18.73\%} & 14.64\% & 16.46\% & 16.44\% & 16.59\% \\
\midrule
\multicolumn{6}{c}{\textbf{Llama3-8B}} \\ \midrule
Question Only (5-shot)       & 7.92\%  & \textbf{9.51\%}  & 8.98\%  & 8.36\%  & 8.69\%  \\
Answer Only (5-shot)         & 23.27\% & 20.87\% & 17.41\% & \textbf{23.56\%} & 21.28\% \\
Question and Answer (5-shot) & 28.16\% & 28.93\% & 23.88\% & \textbf{31.56\%} & 28.63\% \\ 
\midrule
\multicolumn{6}{c}{\textbf{Llama3-70B}} \\ \midrule
Question Only (5-shot)       & 6.75\%  & 6.45\%  & \textbf{9.47\%}  & 5.41\%  & 6.93\%  \\
Answer Only (5-shot)         & 20.85\% & 21.31\% & 18.94\% & \textbf{23.59\%} & 21.31\% \\
Question and Answer (5-shot) & 23.20\% & 24.77\% & 23.10\% & \textbf{27.40\%} & 24.71\% \\ 

\bottomrule
\end{tabular}
\end{table*}

\section{Complete Experiment Results}
\label{sec:complete experiment}
Table~\ref{tab:completeresults} shows our complete experiment results of different LLMs including \texttt{gpt-3.5-turbo}, \texttt{gpt-4o-mini}, \texttt{gemini-1.0-pro}, \texttt{gemini-1.5-pro}, \texttt{llama3-8b} and \texttt{llama3-70b}. The percentage drop of each model is shown in Table~\ref{tab:completedrops}.

\section{Numbers of Human Modified Terms}
\label{sec:Termapp}
\begin{figure*}[t] 
  \centering
  \includegraphics[width=\textwidth]{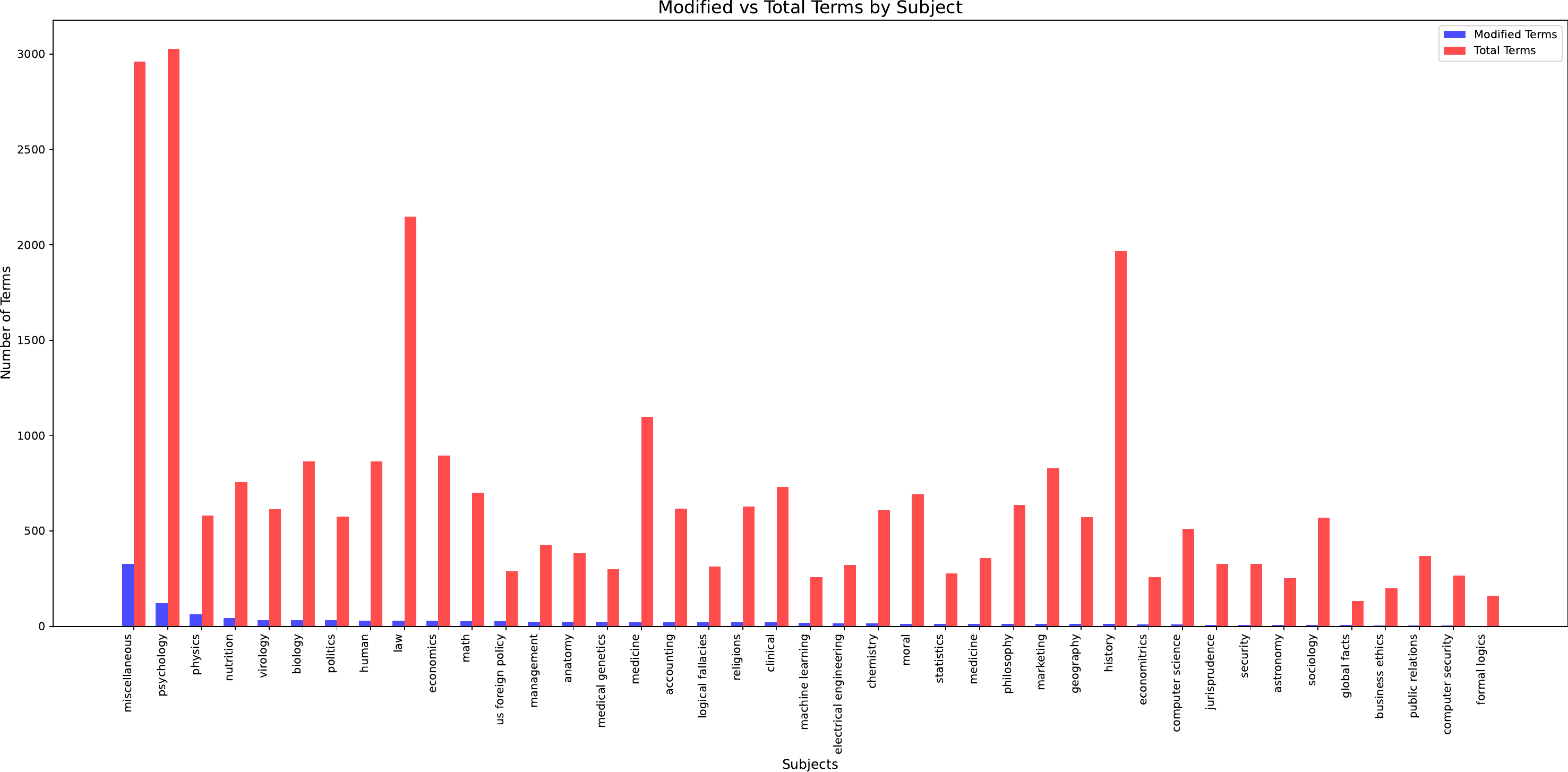} 
  \caption{{Comparison of total generated terms (red) and human-modified terms (blue) across 41 subject glossaries}}
  \label{fig:term}
\end{figure*}

\textcolor{black}{Figure~\ref{fig:term} shows the number of manually modified term definitions across 41 subject glossaries. These glossaries were created by consolidating related topics from the original 57 subjects in the MMLU dataset. Of the 28,676 terms initially generated by \texttt{gpt-3.5-turbo} for these 41 subjects, approximately 1,197 (4.2\%) required human modification.}
\vspace{-0.5cm}

\end{document}